\documentclass[runningheads]{llncs}
\usepackage[T1]{fontenc}
\usepackage{graphicx}
\usepackage{color, colortbl}
\usepackage[breaklinks,colorlinks=true,linkcolor=blue,citecolor=blue,urlcolor=blue]{hyperref}

\usepackage{amsmath}
\usepackage{amssymb}
\usepackage[mathscr]{euscript}
\usepackage{multirow,multicol}

\usepackage{booktabs}
\usepackage{mathtools}
\usepackage{mathabx}
\usepackage{float}
\usepackage[usernames,dvipsnames,svgnames,table]{xcolor}

\usepackage[linesnumbered, vlined, ruled, commentsnumbered]{algorithm2e}
\definecolor{DnCBG}{rgb}{0.9, 0.9, 1.}

\usepackage{breakcites}

\newcommand{\fn}[1]{\footnotesize{#1}}
\newcommand{\green}[1]{\textcolor[RGB]{0,176,80}{#1}}
\newcommand{\gbf}[1]{\green{\bf{\tiny{\fn{(#1)}}}}}

\usepackage[capitalize]{cleveref}
\crefname{section}{Sec.}{Secs.}
\Crefname{section}{Section}{Sections}
\Crefname{table}{Table}{Tables}
\crefname{table}{Tab.}{Tabs.}

\begin{document}

\title{Towards Foundation Models Learned from Anatomy in Medical Imaging via Self-Supervision}

\titlerunning{Towards Foundation Models Learned from Anatomy in Medical Imaging}

\author{Mohammad Reza Hosseinzadeh Taher\inst{1} \and
 Michael B. Gotway\inst{2} \and
Jianming Liang\inst{1}}

\authorrunning{MR. Hosseinzadeh Taher et al. }

\institute{Arizona State University, Tempe, AZ 85281, USA 
\email{\{mhossei2,jianming.liang\}@asu.edu} \and 
 Mayo Clinic, Scottsdale, AZ 85259, USA\\
\email{Gotway.Michael@mayo.edu}}

\maketitle              

\begin{abstract} 
Human anatomy is the foundation of medical imaging and boasts one striking characteristic: its hierarchy in nature, exhibiting two intrinsic properties: (1) \textit{locality}: each anatomical structure is morphologically distinct from the others; and (2) \textit{compositionality}: each anatomical structure is an integrated part of a larger whole. We envision a foundation model for medical imaging that is \textit{consciously} and \textit{purposefully} developed upon this foundation to gain the capability of ``understanding'' human anatomy and to possess the fundamental properties of medical imaging. As our first step in realizing this vision towards foundation models in medical imaging, we devise a novel self-supervised learning (SSL) strategy that exploits the hierarchical nature of human anatomy. Our extensive experiments demonstrate that the SSL pretrained model, derived from our training strategy, not only outperforms state-of-the-art (SOTA) fully/self-supervised baselines but also enhances annotation efficiency, offering potential few-shot segmentation capabilities with performance improvements ranging from 9\% to 30\% for segmentation tasks compared to SSL baselines. This performance is attributed to the significance of \textit{anatomy comprehension} via our learning strategy, which encapsulates the intrinsic attributes of anatomical structures---\textit{locality} and \textit{compositionality}---within the embedding space, yet overlooked in existing SSL methods. All code and pretrained models are available at  \href{https://github.com/JLiangLab/Eden}{GitHub.com/JLiangLab/Eden}.

\end{abstract}

\section{Introduction and related works}
Foundation models~\cite{Bommasani2021Opportunities}, such as GPT-4~\cite{openai2023gpt4} and DALL·E~\cite{Ramesh2021}, pretrained via self-supervised learning (SSL), have revolutionized natural language processing (NLP) and radically transformed vision-language modeling, garnering significant public media attention~\cite{Manjoo2020}. But, despite the development of numerous SSL methods in medical imaging, their success in this domain lags behind their NLP counterparts. What causes these striking differences? We believe that this is because the SSL methods developed for NLP have proven to be powerful in capturing the underlying structures (foundation) of the English language; thus, a number of intrinsic properties of the language emerge naturally, as demonstrated in~\cite{Manning2020}, while the existing SSL methods lack such capabilities to appreciate the foundation of medical imaging---human anatomy. Therefore, this paper is seeking to answer a fundamental question: \textit{How to learn foundation models from human anatomy in medical imaging}?

\begin{figure*}[t]
  \centering
  \includegraphics[width=0.9\linewidth]{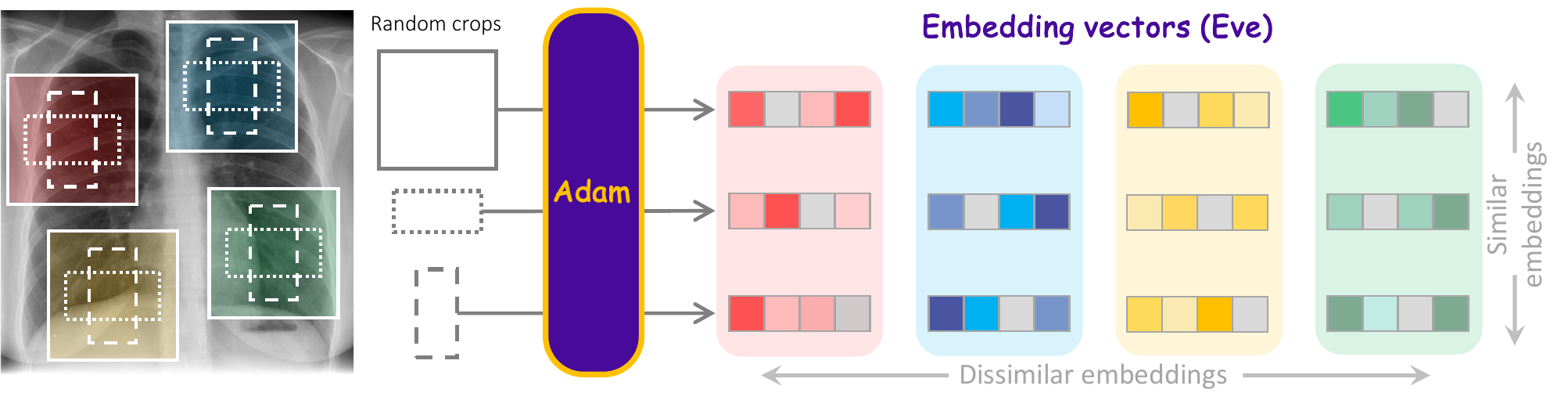}
   \caption{Existing SSL methods lack capabilities of ``understanding'' the foundation of medical imaging---human anatomy. We believe that a foundation model must be able to transform each pixel in an image (e.g., a chest X-ray) into semantics-rich numerical vectors, called embeddings, where different anatomical structures (indicated by different colored boxes) are associated with different embeddings, and the same anatomical structures have (nearly) identical embeddings at all resolutions and scales (indicated by different box shapes) across patients. Inspired by the hierarchical nature of human anatomy (Fig. \textcolor{blue}{6} in Appendix), we introduce a novel SSL strategy to learn anatomy from medical images (\cref{fig:method}), resulting in embeddings (Eve), generated by our pretrained model (Adam), with such desired properties (\cref{fig:local_comp} and Fig. \textcolor{blue}{8} in Appendix).
}   
   \label{fig:intuition}
\end{figure*}

Human anatomy exhibits natural hierarchies. For example, the lungs are divided into the right  and left lung (see Fig. \textcolor{blue}{6} in Appendix) and each lung is further divided into lobes, two on the left and three on the right lung. Each lobe is further subdivided into segments, each containing pulmonary arteries, veins, and bronchi which branch in predictable, dichotomous fashion.  Consequently, anatomical structures have two important properties: \textit{locality}: each anatomical structure is morphologically distinct from others; \textit{compositionality}: each anatomical structure is an integrated part of a larger whole. Naturally, a subquestion is \textit{how to exploit the anatomical hierarchies for training foundation models}? To this end, we devise a novel SSL training strategy, which is \textit{hierarchical}, \textit{autodidactic}, and \textit{coarse}, resulting in a pretrained model, which is \textit{versatile}, and leading to anatomical embedding, which is \textit{dense} and \textit{semantics-meaningful}. Our training strategy is \textit{hierarchical} because it decomposes and perceives the anatomy progressively in a coarse-to-fine manner (\cref{sec:method}\textcolor{blue}{.1}); \textit{autodidactic} because it learns from anatomy through self-supervision, thereby requiring no anatomy labeling (\cref{sec:method}); and \textit{coarse} because it generates dense anatomical embeddings without relying on pixel-level training (\cref{sec:resuts}, ablation 1). The pretrained model is \textit{versatile} because it is strong in generality and adaptability, resulting in performance boosts (\cref{sec:resuts}\textcolor{blue}{.1}) and annotation efficiency (\cref{sec:resuts}\textcolor{blue}{.2}) in myriad tasks. The generated anatomical embedding is \textit{dense} and \textit{semantics-rich} because it possesses two intrinsic properties of anatomical structures, \textit{locality}  (\cref{sec:resuts}\textcolor{blue}{.3}) and \textit{compositionality}  (\cref{sec:resuts}\textcolor{blue}{.4}), in the embedding space, both of which are essential for anatomy understanding. We call our pretrained model \textbf{Adam} (\textit{a}utodidactic \textit{d}ense \textit{a}natomical \textit{m}odels) because it learns autodidactically and yields dense anatomical embedding, nicknamed \textbf{Eve} (\textit{e}mbedding \textit{ve}ctors) for semantic richness (\cref{fig:intuition}). We further coin our project site \textbf{Eden} (\textit{e}nvironment for \textit{d}ense \textit{e}mbeddings and \textit{n}etworks), where all code, pretrained Adam and Eve are placed. 

In summary, we make the following contributions: \textbf{(1)} A novel self-supervised learning strategy that progressively learns anatomy in a coarse-to-fine manner via hierarchical contrastive learning; \textbf{(2)} A new evaluation approach that facilitates analyzing the interpretability of deep models in anatomy understanding by measuring the locality and compositionality of anatomical structures in embedding space;  and \textbf{(3)} A comprehensive and insightful set of experiments that evaluate Adam for a wide range of 9 target tasks, involving fine-tuning, few-shot learning, and investigating semantic richness of Eve in anatomy understanding.

\medskip
\noindent\textbf{Related works:} \noindent\textbf{(i) Self-supervised learning} methods, particularly contrastive techniques~\cite{azizi2021big,Kaku2021Intermediate}, have shown great promise in medical imaging~\cite{Tajbakhsh2021Guest,Taher2021Systematic}. But, due to their focus on image-level features, they are sub-optimal for dense recognition tasks~\cite{Wang2021Dense}. Recent works~\cite{Haghighi2022DiRA,Taher2022CAiD} empower contrastive learning with more discriminative features via using the diversity in the local context of medical images. In contrast to them, which overlook anatomy hierarchies in their learning objectives, Adam exploits the hierarchical nature of anatomy to learn semantics-rich dense features.
\textbf {(ii) Anatomy learning} methods integrate anatomical cues into their SSL objectives. But, GLC~\cite{Chaitanya2020Contrastive} requires spatial correspondence across images, limiting its scalability to non-aligned images. Although TransVW~\cite{haghighi2021transferable}, SAM~\cite{Yan2022SAM}, and Alice~\cite{jiang2023anatomical} relax this requirement, they neglect hierarchical anatomy relations, offering no compositionality. By contrast, Adam learns consistent anatomy features without relying on spatial alignment across images (see Fig. \textcolor{blue}{7} in Appendix) and captures both local and global contexts hierarchically to offer both locality and compositionality. \textbf {(iii) Hierarchical SSL} methods exploit transformers' self-attention to model dependencies among image patches. But, they  fail to capture  anatomy relations due to inefficient SSL signals that 
contrast similar anatomical structures~\cite{Tang2022Self} or disregard relations among images~\cite{Xiao2023Delving,Xie2022SimMIM}. Adam goes beyond architecture design by introducing a learning strategy that decomposes anatomy into a hierarchy of parts for coarse-to-fine anatomy learning, and avoids semantic collision in its supervision signal.

\begin{figure*}[t]
  \centering
  \includegraphics[width=0.9\linewidth]{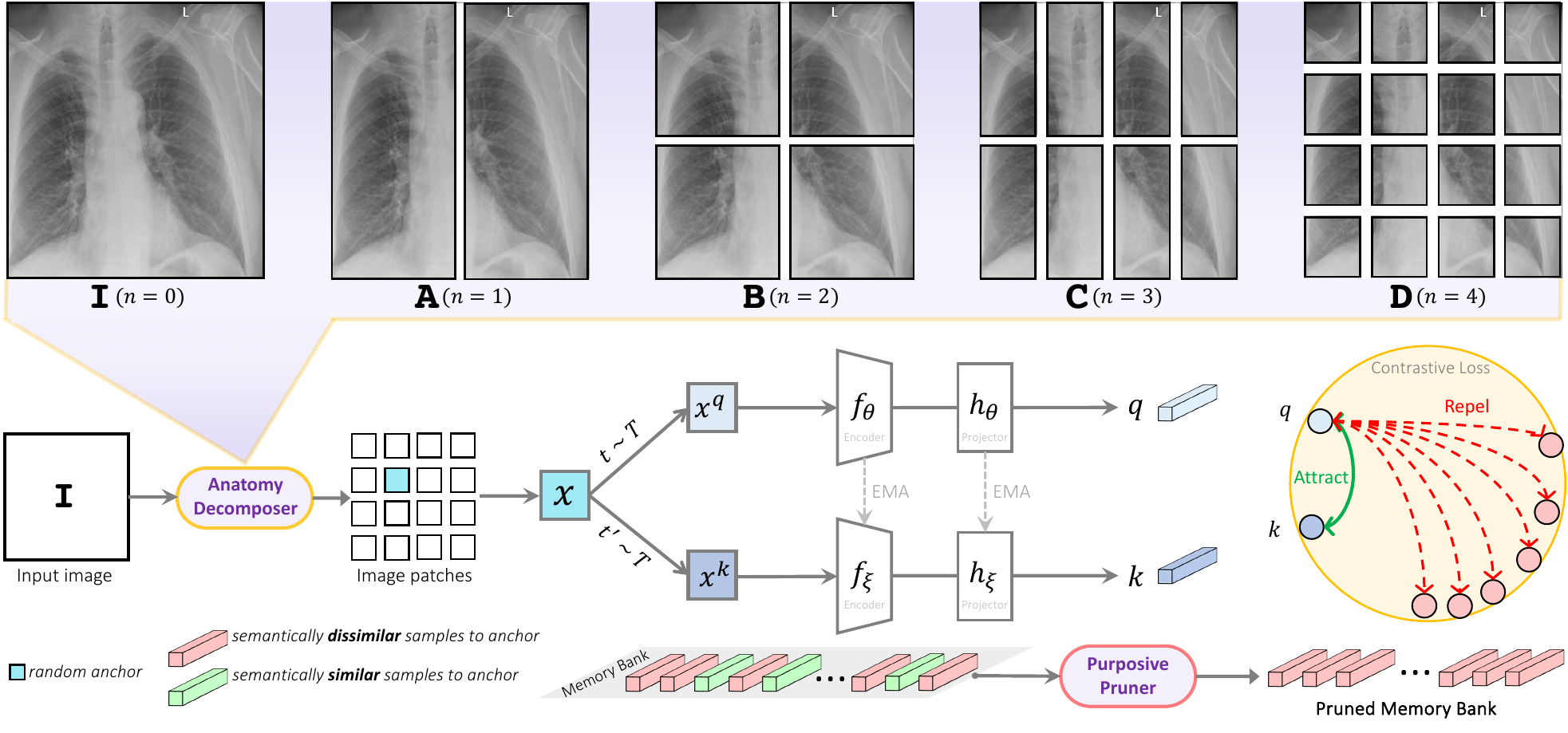}
   \caption{Our SSL strategy gradually decomposes and perceives the anatomy in a coarse-to-fine manner. Our Anatomy Decomposer (AD) decomposes the anatomy into a hierarchy of parts with granularity level $n\in\{0,1,..\}$ at each training stage. Thus, anatomical structures of finer-grained granularity will be incrementally presented to the model as the input. Given image $I$, we pass it to  AD to get a random anchor $x$. We augment $x$ to generate two views (positive samples), and pass them to two encoders to get their  features. To avoid semantic collision in training objective, our Purposive Pruner removes \textit{semantically} similar anatomical structures across images to anchor $x$ from the memory bank. Contrastive loss is then calculated using positive samples' features  and the pruned memory bank. The figure shows pretraining at $n=4$.}   

   \label{fig:method}
\end{figure*}

\section{Method}
\label{sec:method}

Our self-supervised learning strategy, depicted in \cref{fig:method}, aims to exploit the hierarchical nature of human anatomy in order to capture not only generic but also semantically meaningful representations. The main intuition behind our learning strategy is the principle of totality in \textit{Gestalt} psychology: humans commonly first recognize the prominent objects in an image (e.g., lungs) and then \textit{gradually} recognize smaller details based on prior knowledge about that object (e.g., each lung is divided into lobes)~\cite{Sun2020Progressive}. Inspired by this principle, we propose a training strategy, which decomposes and perceives the anatomy progressively in a coarse-to-fine manner, aiming to learn both anatomical (local and global) contextual information and also the relative hierarchical relationships among anatomical structures. Our framework is comprised of two key components: 

\medskip
\noindent \textbf{(1) Anatomy Decomposer (AD)} is responsible for decomposing relevant anatomy into a hierarchy of anatomical structures to guide the model to learn hierarchical anatomical relationships in images. The AD component takes two inputs: an image $I$ and an anatomy granularity level $n$, and generates a random anatomical structure instance $x$. We generate anatomical structures at desired granularity level $n$ in a recursive manner. Given an image $I$, we first split it vertically into two halves (A in \cref{fig:method}). 
Then, we iteratively alternate between horizontally and vertically splitting the resulting image parts until we reach the desired granularity level (B, C, D in \cref{fig:method}). This process results in $2^n$ image patches $\{x_i\}_{i=1}^{2^n}$. In this set, we randomly sample an instance $x$, which will be used as the input for training the model. As such, during the pretraining, anatomical structures at various granular levels are generated and present to the model. 

\medskip
\noindent \textbf{(2) Purposive Pruner (PP)} is responsible for compeling the model to comprehend anatomy more effectively via learning a wider range of distinct anatomical structures. Intuitively, similar anatomical structures (e.g. ribs or disks) should have similar embeddings, while also their finer-grained constituent parts (e.g. different ribs or disks) have (slightly) different embeddings. To achieve such desired embedding space, the anatomical structures need to be intelligently contrasted from each other. Our PP module, in contrast to standard contrastive learning approaches, identifies \textit{semantically} similar anatomical structures in the embedding space and prevents them from being undesirably repelled. In particular, given an anchor anatomical structure $x$ randomly sampled from image $I$, we compute the cosine similarities between features of $x$ and the ones of the points in the  memory bank, and remove the samples with a similarity greater than a threshold $\gamma$ from the  memory bank. Thus, our PP prevents semantic collision, yielding a more optimal embedding space where similar anatomical structures are grouped together while distinguished from dissimilar anatomical structures.

\medskip
\noindent \textbf{Overall training.}
Our framework consists of  two twin backbones  $f_\theta$ and $f_\xi$, and projection heads $h_\theta$ and $h_\xi$. $f_\theta$ and  $h_\theta$ are updated by back-propagation, while  $f_\xi$ and $h_\xi$ are updated by exponential moving average (EMA) of $f_\theta$ and $h_\theta$ parameters, respectively. We use a memory bank to store the embeddings of negative samples $\mathrm{MB}=\{k_i\}_{i=1}^K$, where $K$ is the memory bank size. For learning anatomy in a coarse-to-fine manner, we  progressively increase the anatomical structures granularity. Thus, at each training stage, anatomical structures with granularity level $n\in \{0,1,..\}$ will be presented to the model. Given input image $I$ and data granularity level $n$, we pass them to our AD to get a random anatomical structure $x$. We apply an augmentation function $T(.)$ on $x$ to generate two  views $x_q$ and $x_k$, which are then processed by backbones and projection heads to generate latent features $q=h_\theta(f_\theta(x_q))$ and $k=h_\xi(f_\xi(x_k))$. Then, we pass $q$ and $\mathrm{MB}$ to our PP to remove false negative samples for anchor $x$, resulting in pruned memory bank $\mathrm{MB_{pruned}}$, which is used to compute the InfoNCE~\cite{chen2020improved} loss $\mathcal{L} = - log \frac{exp(q\cdot k/\tau)}{exp(q\cdot k/\tau)+\sum\limits_{i=1}^{K'} exp(q \cdot k_i/\tau)}$, where $\tau$ is a temperature hyperparameter, $K'$ is size of $\mathrm{MB_{pruned}}$, and $k_i \in \mathrm{MB_{pruned}}$. Our AD module enables the model to first learn anatomy at a coarser-grained level, and then use this acquired knowledge as effective  contextual clues for learning more fine-grained anatomical structures, reflecting anatomical structures compositionality in its embedding space. Our PP module enables the model to learn a semantically-structured embedding space that preserves anatomical structures locality by removing semantic collision from the model's learning objective. The pretrained model derived by our training strategy (\textbf{Adam}) can not only be used as a basis for myriad target tasks via adaptation (\textit{fine-tuning}), but also its embedding vectors (\textbf{Eve}) show promises to be used standalone \textit{without} adaptation for other tasks like landmark detection.

\section{Experiments and Results}
\label{sec:resuts}

\begin{figure}[t]
  \centering
  \includegraphics[width=1\linewidth]{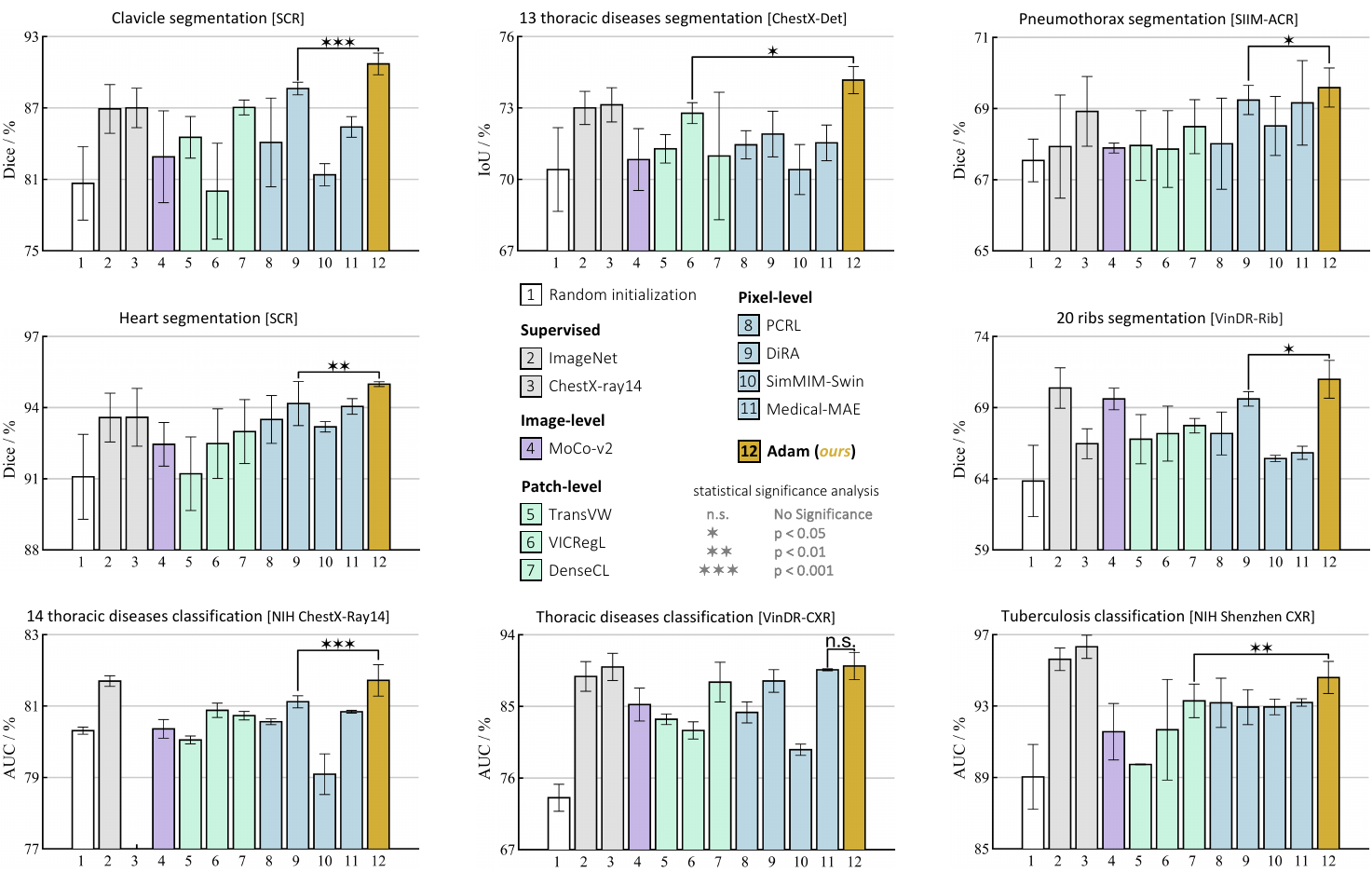}
   \caption{Adam provides superior performance over fully/self-supervised methods. All SSL methods are pretrained on ChestX-ray14 dataset. Statistical significance analysis ($p < 0.05$) was conducted between Adam and the top SSL baseline in each task.}   

   \label{fig:transfer_sota}
\end{figure}

\noindent \textit{\textbf{Pretraining and fine-tuning settings:}} We use unlabeled training images of ChestX-ray14~\cite{wang2017chestx} and EyePACS~\cite{Cuadros2009EyePACS} for pretraining and follow \cite{chen2020improved} in pretraining settings: SGD optimizer with an initial learning rate of 0.03, weight decay 1e-4, SGD momentum 0.9, cosine decaying scheduler, and batch size 256. The input anatomical structures are resized to 224$\times$224; augmentations include random crop, color jitter, Gaussian blur, and rotation. We use data granularity level ($n$) up to 4 and pruning threshold $\gamma=0.8$ (ablation in Appendix). Following~\cite{Kaku2021Intermediate,Haghighi2022DiRA}, we adopt ResNet-50 as the backbone. For fine-tuning, we (1) use the pretrained encoder followed by a task-specific head for classification tasks, and a U-Net network for segmentation tasks where the encoder is initialized with the pretrained backbone; (2) fine-tune all downstream model’s params; (3) run each method 10 times on each task and report statistical significance analysis. 

\medskip
\noindent\textit{\textbf{Downstream tasks and baselines:}} We evaluate Adam on a myraid of 9 tasks on ChestX-ray14~\cite{wang2017chestx}, Shenzhen~\cite{Jaeger2014Tow}, VinDr-CXR~\cite{nguyen2020vindrcxr}, VinDR-Rib~\cite{Nguyen2021VinDr}, SIIM-ACR~\cite{PNEchallenge}, SCR~\cite{vanginneken2006Segmentation}, ChestX-Det~\cite{Lian2021Structure}, and DRIVE~\cite{Budai2013Robust}, covering various challenging tasks, diseases, and organs. We compare Adam with SOTA \textit{image-} (MoCo-v2~\cite{chen2020improved}), \textit{patch-} (TransVW~\cite{haghighi2021transferable}, VICRegL~\cite{Bardes2022VICRegL}, DenseCL~\cite{Wang2021Dense}),
and \textit{pixel-level} (PCRL~\cite{Zhou2021Preservational}, DiRA~\cite{Haghighi2022DiRA}, Medical-MAE~\cite{Xiao2023Delving}, SimMIM~\cite{Xie2022SimMIM}) SSL methods.

\medskip
\noindent\textbf{1) Adam provides generalizable representations for a variety of tasks.} To showcase the significance of anatomy learning via our SSL approach and its impact on representation learning, we compare transfer learning performance of Adam to 8 recent SOTA SSL methods with diverse objectives, as well as 2 fully-supervised models pretrained on ImageNet and ChestX-ray14 datasets, in 8 downstream tasks. As seen in \cref{fig:transfer_sota}, (\textit{i}) our Adam consistently outperforms the SOTA dense SSL methods (VICRegL \& DenseCL) as well as the SOTA medical SSL methods (PCRL \& DiRA), and achieves superior or comparable performance compared to fully-supervised baselines; (\textit{ii}) our Adam demonstrates a significant performance improvement over TransVW, which is specifically designed for learning recurring anatomical structures across patients. This emphasizes the effectiveness of our coarse-to-fine approach in capturing both local and global context of anatomical structures hierarchically, in contrast to TransVW which learns them at a fixed level; and (\textit{iii}) our Adam remains superior to ViT-based SSL methods such as Medical-MAE and SimMIM, which divide the input image into smaller patches and utilize self-attention to model patch dependencies. This underscores the importance of our learning strategy in effectively modeling the hierarchical relationships among anatomical structures.

\begin{table}[t]
\setlength{\tabcolsep}{4.8pt}
   \caption{Few-shot transfer on two medical segmentation tasks. Adam provides outstandingly better performance compared with SSL baselines. \textbf{\textcolor[RGB]{0,176,80}{Green numbers}} show Adam's performance boosts compared with the second-best method in each task/shot.}
   \label{tab:transfer_fewshot}
  \centering
  \scalebox{0.65}{
  \begin{tabular}{l|llll|llll}
\toprule
 \multirow{2}{*}{ Method}& \multicolumn{4}{c|}{SCR-Heart [Dice(\%)]} & \multicolumn{4}{c}{SCR-Clavicle [Dice(\%)]} \\
  \cline{2-5} \cline{6-9}
  & 3-shot & 6-shot & 12-shot & 24-shot &  3-shot & 6-shot & 12-shot & 24-shot \\
 
\toprule
 MoCo-v2 &  44.84&	59.97&	69.90	&79.69 & 23.77&	29.24&	38.07&	44.47 \\

\midrule
  DenseCL & \underline{64.8}8&	\underline{74.43}&	75.79	&80.06& \underline{36.43}&	\underline{51.31}&	63.03&	69.13\\
  
 \midrule
DiRA & 63.76&	64.47&	\underline{76.10}&	\underline{81.42}& 31.42&	38.59&	\underline{66.81}&	\underline{73.06}\\

  \midrule
Adam (ours)  & \textbf{84.35}\gbf{$\uparrow$19}& \textbf{86.70}\gbf{$\uparrow$12}& \textbf{89.79}\gbf{$\uparrow$14}&	\textbf{90.45}\gbf{$\uparrow$9}& \textbf{66.69}\gbf{$\uparrow$30}&	\textbf{79.41}\gbf{$\uparrow$28}&	\textbf{83.96}\gbf{$\uparrow$17}&	\textbf{84.76}\gbf{$\uparrow$12}\\
    \bottomrule
  \end{tabular}
  }
 	
\end{table}

\medskip
\noindent\textbf{2) Adam enhances annotation efficiency, revealing promise for few-shot learning.} To dissect robustness of our representations, we compare Adam with top-performing SSL methods from each baseline group, based on \cref{fig:transfer_sota}, in limited data regimes. We conduct experiments on Heart and Clavicle segmentation tasks, and fine-tune the pretrained models using a few shots of labeled data (3, 6, 12, and 24) randomly sampled from each dataset. As seen in \cref{tab:transfer_fewshot}, Adam not only demonstrates superior performance against baselines by a large margin (\textbf{\textcolor[RGB]{0,176,80}{Green \textit{nums.}}}) but also maintains consistent behavior with minimal performance drop as labeled data decreases, compared to baselines. We attribute Adam’s superior representations over baselines, as seen in \cref{fig:transfer_sota} and \cref{tab:transfer_fewshot}, to its ability to learn the anatomy by preserving locality and compositionality of anatomical structures in its embedding space, as is exemplified in the following.

\begin{figure*}[t]
  \centering
  \includegraphics[width=1\linewidth]{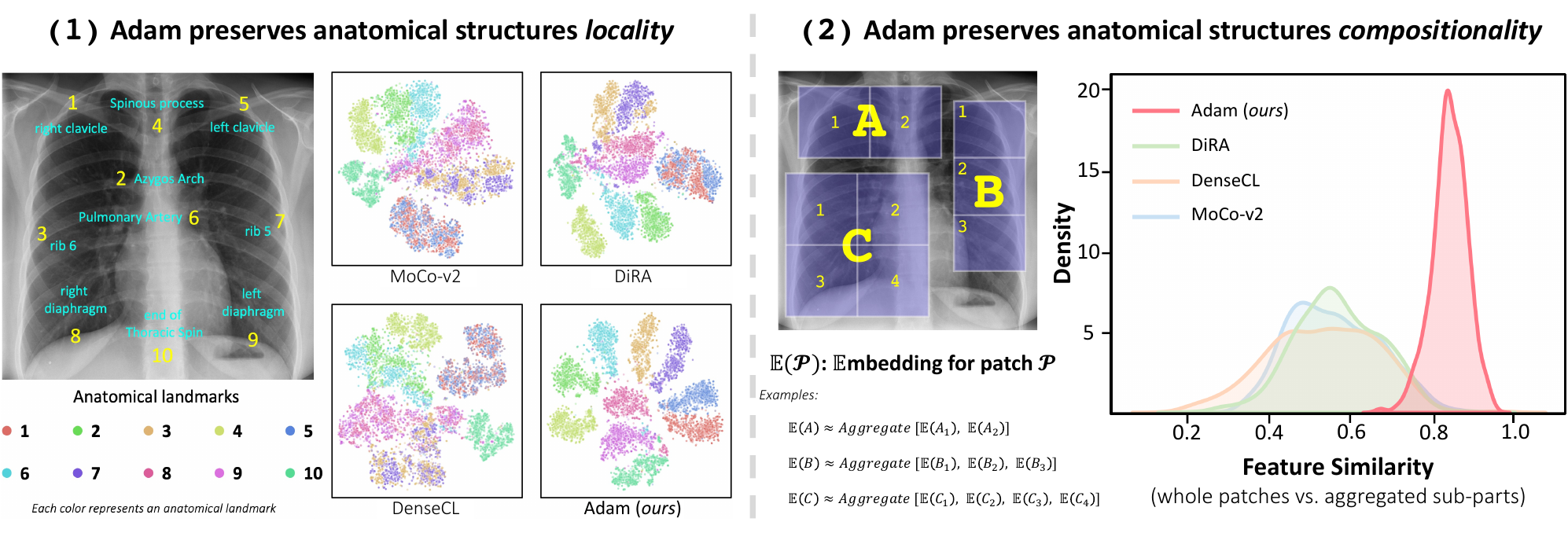}
   \caption{Adam preserves locality and compositionality properties, which are intrinsic to anatomical structures and critical for understanding anatomy, in its embedding space.}  
   \label{fig:local_comp}
\end{figure*}

\medskip
\noindent\textbf{3) Adam preserves anatomical structures locality.} We investigate Adam's ability to reflect \textit{locality} of anatomical structures in its embedding space against existing SSL baselines. To do so, we (1) create a dataset of 1,000 images (from ChestX-ray14 dataset) with 10 distinct anatomical landmarks \textit{manually} annotated by human experts in each image, (2) extract 224$\times$224 patches around each landmark across images, (3) extract latent features of each landmark instance using each pretrained model under study and then pass them through a global average pooling layer, and (4) visualize the features by using t-SNE. As seen in \cref{fig:local_comp}\textcolor{blue}{.1}, existing SSL methods lack the ability in discriminating different anatomical structures, causing ambiguous embedding spaces. In contrast, Adam excels in distinguishing various anatomical landmarks, yielding well-separated clusters in its embedding space. This highlights Adam's ability to learn a rich semantic embedding space where distinct anatomical structures have unique embeddings, and identical structures share near-identical embeddings across patients.

\medskip
\noindent\textbf{4) Adam preserves anatomical structures compositionality.} The embedding of a whole should be equal or close to the sum of the embedding of its each part (see $\mathbb{E}$($\mathscr{P}$) examples in \cref{fig:local_comp}\textcolor{blue}{.2}). To investigate Adam's ability to reflect \textit{compositionality} of anatomical structures in its embedding space against existing SSL baselines, we (1) extract random patches from test images of ChestX-ray14, and decompose each patch into 2,3, or 4 non-overlapping sub-patches, (2) resize each extracted patch and its sub-patches to 224$\times$224 and then extract their features using each pretrained model under study, (3) compute cosine similarity between the embedding of each patch and the aggregate of the embeddings of its sub-patches, and (4) visualize the similarity distributions with Gaussian kernel density estimation (KDE). As seen in \cref{fig:local_comp}\textcolor{blue}{.2}, Adam's distribution is not only narrower and taller than baselines, but also the mean of similarity value between embedding of whole patches and their aggregated sub-parts is closer to 1.

\begin{figure*}[t]
  \centering
  \includegraphics[width=1\linewidth]{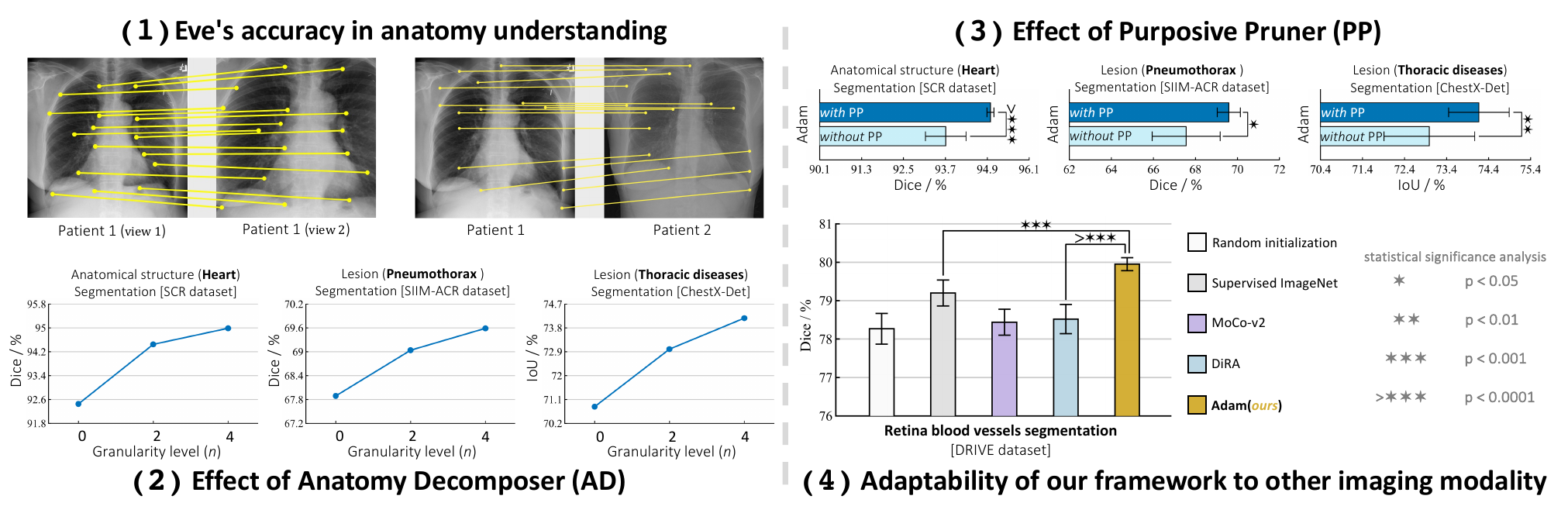}
   \caption{Ablation studies on (1) Eve’s accuracy in anatomy understanding, (2) effect of anatomy decomposer, (3) effect of purposive pruner, and (4) adaptability of our framework to other imaging modalities.}  
   \label{fig:ablation}
\end{figure*}

\medskip
\noindent \textbf{Ablation 1: Eve's accuracy in anatomy understanding} was studied by visualizing dense correspondence between (i) an image and its augmented views and (ii) different images. Given two images, we divide them into grids of patches and extract their features $Eve_1$ and $Eve_2$ using Adam's pretrained model. For each feature vector in $Eve_1$, we find its correspondence in $Eve_2$ based on highest \textit{cosine} similarity; for clarity, we show some of the high-similarity matches ($\geq$0.8) in \cref{fig:ablation}\textcolor{blue}{.1}. As seen, Eve has accurate dense anatomical representations, mapping semantically similar structures, regardless of their differences. Although Adam is not explicitly trained for this purpose, these results show its potential for landmark detection and image registration applications, as an emergent property.

\medskip
\noindent\textbf{Ablation 2: Effect of Anatomy Decomposer} was studied by gradually increasing pretraining data granularity from coarse-grained anatomy ($n=0$) to finer levels (up to $n=4$) and fine-tuning the models on downstream tasks. As seen in \cref{fig:ablation}\textcolor{blue}{.2}, gradual increment of data granularity consistently improves the  performance across all tasks. This suggests that our coarse-to-fine learning strategy deepens the model's anatomical knowledge.

\medskip
\noindent\textbf{Ablation 3: Effect of Purposive Pruner} was studied by comparing a model with and \textit{without} PP (i.e. contrasting an anchor with all negative pairs in the memory bank) during pretraining. \cref{fig:ablation}\textcolor{blue}{.3} shows PP leads to significant performance boosts across all tasks, highlighting its key role in enabling the model to capture more discriminative features by removing noisy contrastive pairs.

\medskip
\noindent\textbf{Ablation 4: Adaptability of our framework to other imaging modalities} was explored by utilizing fundoscopy photography images in EyePACS as pretraining data, which possess complex structures due to the diverse variations in retinal anatomy. As depicted in \cref{fig:ablation}\textcolor{blue}{.4}, Adam provides superior performance by 1.4\% ($p<0.05$) in the blood vessel segmentation task compared to the top-performing SSL methods that also leverage the same pretraining images. This highlights the importance of effectively learning the anatomy and also showcases the potential applicability of our method to various imaging modalities.

\section{{Conclusion and Future Work}}
A key contribution of ours lies in crafting a novel SSL strategy that underpins the development of powerful self-supervised models foundational to medical imaging via learning anatomy. Our training strategy progressively learns anatomy in a coarse-to-fine manner via hierarchical contrastive learning. Our approach yields highly generalizable pretrained models and anatomical embeddings with essential properties of \textit{locality} and \textit{compositionality}, making them semantically meaningful for anatomy understanding. In future, we plan to apply our strategy to provide \textit{dense anatomical models} for major imaging modalities and protocols.


\begin{thebibliography}{10}
\providecommand{\url}[1]{\texttt{#1}}
\providecommand{\urlprefix}{URL }
\providecommand{\doi}[1]{https://doi.org/#1}

\bibitem{PNEchallenge}
Siim-acr pneumothorax segmentation (2019),
  \url{https://www.kaggle.com/c/siim-acr-pneumothorax-segmentation/}

\bibitem{azizi2021big}
Azizi, S., Mustafa, B., Ryan, F., Beaver, Z., Freyberg, J., Deaton, J., Loh,
  A., Karthikesalingam, A., Kornblith, S., Chen, T., Natarajan, V., Norouzi,
  M.: Big self-supervised models advance medical image classification. In:
  Proceedings of the IEEE/CVF International Conference on Computer Vision. pp.
  3478--3488 (2021)

\bibitem{Bardes2022VICRegL}
Bardes, A., Ponce, J., LeCun, Y.: Vicregl: Self-supervised learning of local
  visual features. In: Advances in Neural Information Processing Systems.
  vol.~35, pp. 8799--8810 (2022)

\bibitem{Bommasani2021Opportunities}
Bommasani, R., et~al.: On the opportunities and risks of foundation models.
  ArXiv  (2021), \url{https://crfm.stanford.edu/assets/report.pdf}

\bibitem{Budai2013Robust}
Budai, A., Bock, R., Maier, A., Hornegger, J., Michelson, G.: Robust vessel
  segmentation in fundus images. International Journal of Biomedical Imaging
  (2013)

\bibitem{Chaitanya2020Contrastive}
Chaitanya, K., Erdil, E., Karani, N., Konukoglu, E.: Contrastive learning of
  global and local features for medical image segmentation with limited
  annotations. In: Advances in Neural Information Processing Systems. vol.~33,
  pp. 12546--12558 (2020)

\bibitem{chen2020improved}
Chen, X., Fan, H., Girshick, R., He, K.: Improved baselines with momentum
  contrastive learning (2020)

\bibitem{Cuadros2009EyePACS}
Cuadros, J., Bresnick, G.: Eyepacs: An adaptable telemedicine system for
  diabetic retinopathy screening. Diabetes Science and Technology
  \textbf{3}(3),  509--516 (2009)

\bibitem{vanginneken2006Segmentation}
van Ginneken, B., Stegmann, M., Loog, M.: {Segmentation of anatomical
  structures in chest radiographs using supervised methods: a comparative study
  on a public database}. {Medical Image Analysis}  \textbf{10}(1),  19--40
  (2006)

\bibitem{Haghighi2022DiRA}
Haghighi, F., Hosseinzadeh~Taher, M.R., Gotway, M.B., Liang, J.: Dira:
  Discriminative, restorative, and adversarial learning for self-supervised
  medical image analysis. In: Proceedings of the IEEE/CVF Conference on
  Computer Vision and Pattern Recognition (CVPR). pp. 20824--20834 (2022)

\bibitem{haghighi2021transferable}
Haghighi, F., Taher, M.R.H., Zhou, Z., Gotway, M.B., Liang, J.: Transferable
  visual words: Exploiting the semantics of anatomical patterns for
  self-supervised learning. IEEE Transactions on Medical Imaging
  \textbf{40}(10),  2857--2868 (2021)

\bibitem{Taher2021Systematic}
Hosseinzadeh~Taher, M.R., Haghighi, F., Feng, R., Gotway, M.B., Liang, J.: A
  systematic benchmarking analysis of transfer learning for medical image
  analysis. In: Domain Adaptation and Representation Transfer, and Affordable
  Healthcare and AI for Resource Diverse Global Health. pp. 3--13 (2021)

\bibitem{Taher2022CAiD}
Hosseinzadeh~Taher, M.R., Haghighi, F., Gotway, M.B., Liang, J.: Caid:
  Context-aware instance discrimination for self-supervised learning in medical
  imaging. In: Proceedings of The 5th International Conference on Medical
  Imaging with Deep Learning. Proceedings of Machine Learning Research,
  vol.~172, pp. 535--551 (2022)

\bibitem{Jaeger2014Tow}
Jaeger, S., Candemir, S., Antani, S., Wáng, Y.X.J., Lu, P.X., Thoma, G.: Two
  public chest x-ray datasets for computer-aided screening of pulmonary
  diseases. Quantitative imaging in medicine and surgery  \textbf{4}(6) (2014)

\bibitem{jiang2023anatomical}
Jiang, Y., Sun, M., Guo, H., Yan, K., Lu, L., Xu, M.: Anatomical invariance
  modeling and semantic alignment for self-supervised learning in 3d medical
  image segmentation. arXiv  (2023)

\bibitem{Kaku2021Intermediate}
Kaku, A., Upadhya, S., Razavian, N.: Intermediate layers matter in momentum
  contrastive self supervised learning. In: Advances in Neural Information
  Processing Systems. pp. 24063--24074 (2021)

\bibitem{Lian2021Structure}
Lian, J., Liu, J., Zhang, S., Gao, K., Liu, X., Zhang, D., Yu, Y.: A
  structure-aware relation network for thoracic diseases detection and
  segmentation. IEEE Transactions on Medical Imaging  \textbf{40}(8),
  2042--2052 (2021)

\bibitem{Manjoo2020}
Manjoo, F.: {How Do You Know a Human Wrote This}. The New York Times  (2020)

\bibitem{Manning2020}
Manning, C.D., Clark, K., Hewitt, J., Khandelwal, U., Levy, O.: {Emergent
  linguistic structure in artificial neural networks trained by
  self-supervision}. Proceedings of the National Academy of Sciences
  \textbf{117}(48),  30046--30054 (2020)

\bibitem{nguyen2020vindrcxr}
Nguyen, H.Q., Lam, K., Le, L.T., , et~al.: Vindr-cxr: An open dataset of chest
  x-rays with radiologist's annotations. Scientific Data  \textbf{9}, ~429
  (2020)

\bibitem{Nguyen2021VinDr}
Nguyen, H.C., Le, T.T., Pham, H.H., Nguyen, H.Q.: Vindr-ribcxr: A benchmark
  dataset for automatic segmentation and labeling of individual ribs on chest
  x-rays. In: Medical Imaging with Deep Learning (2021)

\bibitem{openai2023gpt4}
OpenAI: Gpt-4 technical report (2023)

\bibitem{Ramesh2021}
Ramesh, A., Pavlov, M., Goh, G., Gray, S., Voss, C., Radford, A., Chen, M.,
  Sutskever, I.: Zero-shot text-to-image generation. In: Proceedings of the
  38th International Conference on Machine Learning. vol.~139, pp. 8821--8831
  (2021)

\bibitem{Sun2020Progressive}
Sun, Y., Hu, J., Shi, J., Sun, Z.: Progressive decomposition: A method of
  coarse-to-fine image parsing using stacked networks. Multimedia Tools Appl.
  \textbf{79}(19–20),  13379–13402 (2020)

\bibitem{Tajbakhsh2021Guest}
Tajbakhsh, N., Roth, H., Terzopoulos, D., Liang, J.: Guest editorial
  annotation-efficient deep learning: The holy grail of medical imaging. IEEE
  Transactions on Medical Imaging  \textbf{40}(10),  2526--2533 (2021)

\bibitem{Tang2022Self}
Tang, Y., Yang, D., Li, W., Roth, H.R., Landman, B., Xu, D., Nath, V.,
  Hatamizadeh, A.: Self-supervised pre-training of swin transformers for 3d
  medical image analysis. In: Proceedings of the IEEE/CVF Conference on
  Computer Vision and Pattern Recognition (CVPR). pp. 20730--20740 (2022)

\bibitem{wang2017chestx}
Wang, X., Peng, Y., Lu, L., Lu, Z., Bagheri, M., et~al.: Chestx-ray8:
  Hospital-scale chest x-ray database and benchmarks on weakly-supervised
  classification and localization of common thorax diseases. In: Proceedings of
  the IEEE/CVF Conference on Computer Vision and Pattern Recognition (CVPR).
  pp. 2097--2106 (2017)

\bibitem{Wang2021Dense}
Wang, X., Zhang, R., Shen, C., Kong, T., Li, L.: Dense contrastive learning for
  self-supervised visual pre-training. In: Proceedings of the IEEE/CVF
  Conference on Computer Vision and Pattern Recognition (CVPR). pp. 3024--3033
  (2021)

\bibitem{Xiao2023Delving}
Xiao, J., Bai, Y., Yuille, A., Zhou, Z.: Delving into masked autoencoders for
  multi-label thorax disease classification. In: Proceedings of the IEEE/CVF
  Winter Conference on Applications of Computer Vision (WACV). pp. 3588--3600
  (2023)

\bibitem{Xie2022SimMIM}
Xie, Z., Zhang, Z., Cao, Y., Lin, Y., Bao, J., Yao, Z., Dai, Q., Hu, H.:
  Simmim: A simple framework for masked image modeling. In: Proceedings of the
  IEEE/CVF Conference on Computer Vision and Pattern Recognition. pp.
  9653--9663 (2022)

\bibitem{Yan2022SAM}
Yan, K., Cai, J., Jin, D., Miao, S., Guo, D., Harrison, A.P., Tang, Y., Xiao,
  J., Lu, J., Lu, L.: Sam: Self-supervised learning of pixel-wise anatomical
  embeddings in radiological images. IEEE Transactions on Medical Imaging
  \textbf{41}(10),  2658--2669 (2022)

\bibitem{Zhou2021Preservational}
Zhou, H.Y., Lu, C., Yang, S., Han, X., Yu, Y.: Preservational learning improves
  self-supervised medical image models by reconstructing diverse contexts. In:
  Proceedings of the IEEE/CVF International Conference on Computer Vision
  (ICCV). pp. 3499--3509 (2021)

\end{thebibliography}

\title{---Supplementary Material--- \\ Towards Foundation Models Learned from Anatomy in Medical Imaging via Self-Supervision}

\author{Mohammad Reza Hosseinzadeh Taher\inst{1} \and
 Michael B. Gotway\inst{2} \and
Jianming Liang\inst{1}}


\authorrunning{MR. Hosseinzadeh Taher et al. }
\titlerunning{Towards Foundation Models Learned from Anatomy in Medical Imaging}

%
\institute{Arizona State University, Tempe, AZ 85281, USA 
\email{\{mhossei2,jianming.liang\}@asu.edu} \and 
 Mayo Clinic, Scottsdale, AZ 85259, USA\\
\email{Gotway.Michael@mayo.edu}}
\maketitle  

\setcounter{figure}{5}
\appendix

\section{The intuition behind our proposed SSL framework}
\begin{figure*}[!h]
  \centering
  \includegraphics[width=0.62\linewidth]{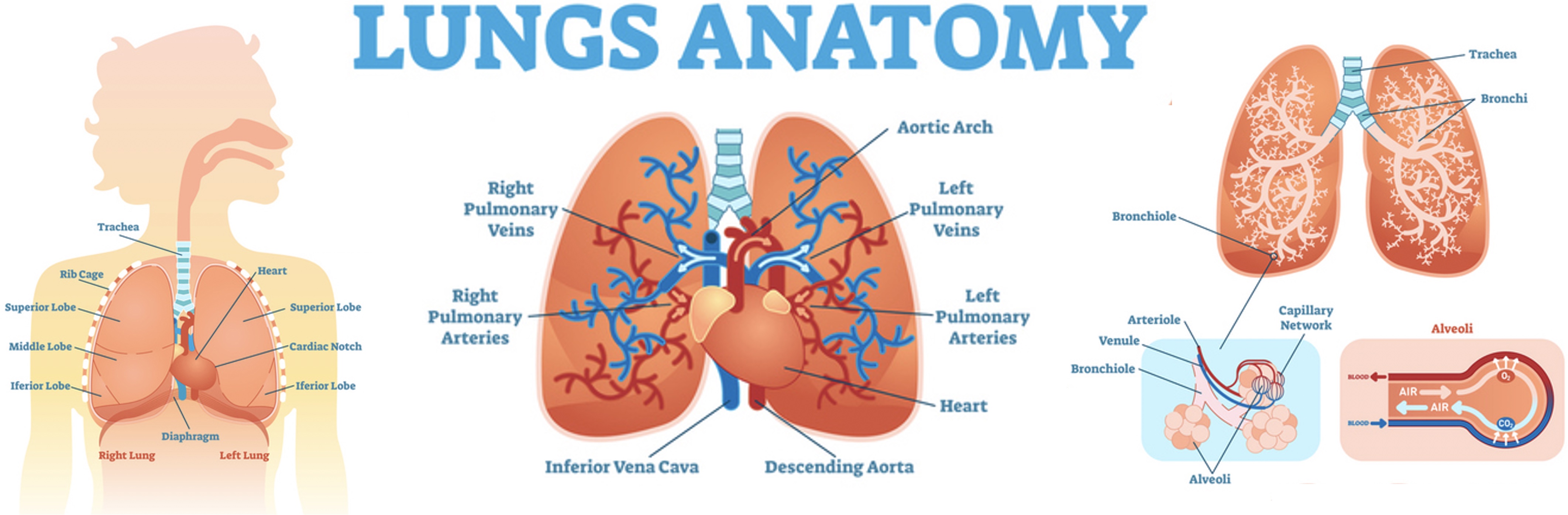}
   \caption{Human anatomy exhibit natural hierarchies. For example, lung divided into right and left lung, each with lobes. The right lung has three lobes: superior, middle, and inferior; the left lung has two lobes: superior and inferior. The pulmonary arteries, veins, and airways form hierarchical trees. These anatomy hierarchies have inspired us to propose a SSL strategy that captures locality and compositionality  of anatomical structures in its embedding space, crucial for anatomy understanding, yet overlooked in existing SSL methods. The image for the lung anatomy available at https://stock.adobe.com/.}  
   \label{fig:lung_anatomy}
\end{figure*}

\begin{figure}[!h]
  \centering
  \includegraphics[width=1\linewidth]{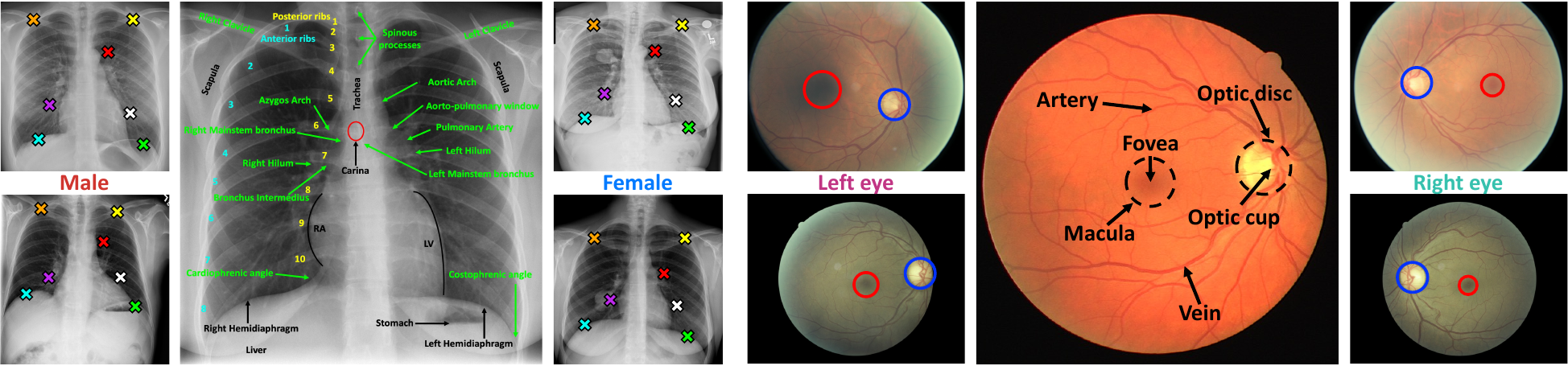}
   \caption{The anatomical similarity of medical images generated from a particular imaging protocol yields consistent hierarchical anatomical structures, which can be placed at different spatial locations across images due to inter-subject variations. This paper exploits the intrinsic anatomical hierarchies in medical images for SSL, yielding consistent anatomical embeddings without relying on \textit{spatial} correspondence across patients. }   
   \label{fig:illustration_appendix}
\end{figure}

\begin{figure}[!h]
  \centering
  \includegraphics[width=0.9\linewidth]{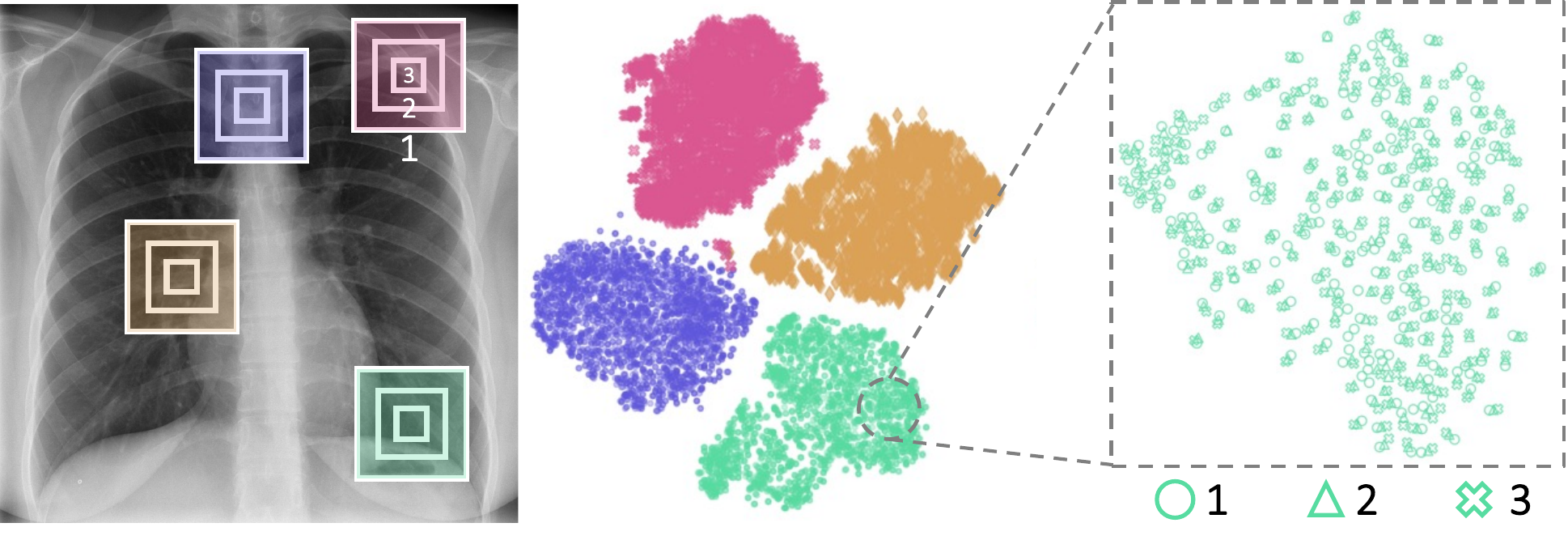}
  \caption{Adam is capable of generating semantics-rich dense embeddings (Eve), where different anatomical structures are associated with different embeddings, and the same anatomical structures have (nearly) identical embeddings at all resolutions and scales. 
  }
 \label{fig:comp_intuition}
\end{figure}

\section{Adam's capability in anatomy understanding}
We delve deeper into Adam’s capability to generate semantics-rich dense embeddings, where different anatomical structures are associated with different embeddings, and the same anatomical structures have (nearly) identical embeddings at all resolutions and scales.  To do so, we employ a dataset comprising 1,000 images along with 4 distinct anatomical landmarks annotated  in each image (details in Sec. 3.3). We then extract three patches of different resolutions, denoted as levels 1, 2, and 3, around each landmark location across the images. As a result, instances of each of the four distinct anatomical landmarks represent different anatomical structures. Furthermore, the anatomical structures corresponding to these four landmarks at level 1 exhibit close similarity to their corresponding structures at levels 2 and 3.  All anatomical structures in each level are resized to  224$\times$224, and Adam’s pretrained model is used to extract their embeddings (i.e. Eve). Finally, tSNE was used to visualize the embeddings. As seen in \cref{fig:comp_intuition}, the instances of four distinct anatomical landmarks (represented by four different colors) are well-separated from one another, highlighting Adam's capability in distinguishing different anatomical structures.  Moreover, the embeddings of the anatomical structures at levels 1, 2, and 3 for each of the four landmarks are close to each other, echoing Adam’s ability to provide (almost) identical embeddings for similar anatomical structures across different resolutions.

\section{Additional results}
\begin{figure}[t]
  \centering
  \includegraphics[width=0.9\linewidth]{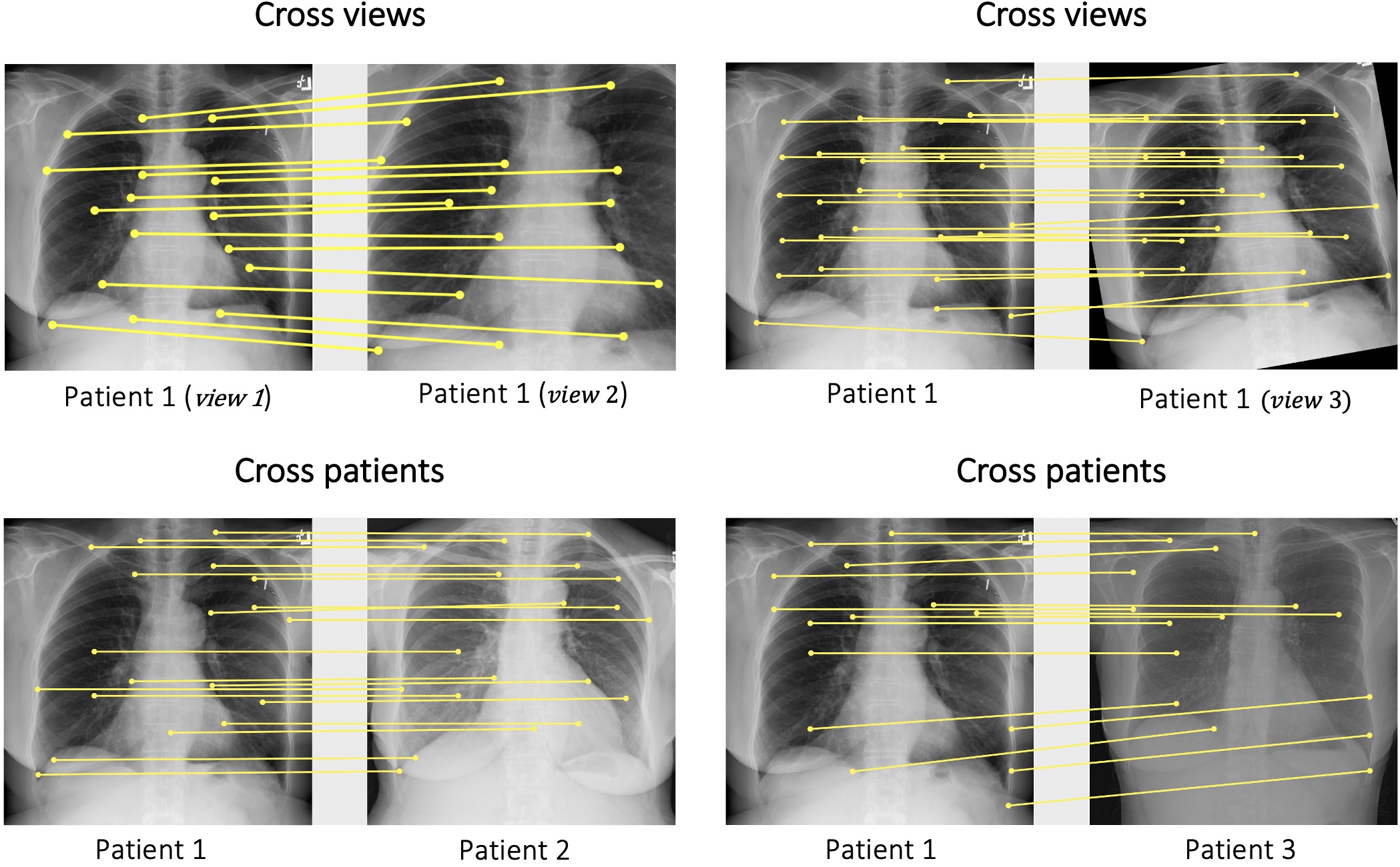}
  \caption{Visualization of dense correspondence  provided by Eve across different views of the same image (first row) and different patients with diversity in intensity distribution and organs' appearance (second row).}
 \label{fig:appendix_dense_correspondence}
\end{figure}

\subsection{Dense correspondence visualization}
To further demonstrate the Eve’s accuracy in anatomy understanding, we explore the Eve's robustness to (i) image augmentations and (ii) variations in appearance, intensity, and texture of anatomical structures caused by inter-subject differences or data distribution shifts. To do so, we visualize the dense correspondence between (i) an image and its augmented views produced by cropping and rotation (10 degrees) and (ii) images of different patients with considerable diversity in intensity distribution, texture, and organs' shape. For clarity of figures, we only show some of the high-similarity matches. A match between two feature vectors is represented by a yellow line. \cref{fig:appendix_dense_correspondence} shows Eve is capable of finding similar anatomical patterns across the different views or even across patients. We conclude that Eve provides accurate anatomical representations, mapping semantically similar anatomical structures, regardless of their subtle differences in shape, intensity, and texture, to similar embeddings. Although our method is not designed for this purpose, these results show its potential for landmark detection and image registration applications. It should be noted that our method's primary goal is to provide generalizable models; thus, while our Eve shows some potential for dense visual correspondence, more detailed investigation and comparisons with SOTA methods in this context, such as~\cite{Yan2022SAM}, are required, which we leave to future work.

\begin{figure}[t]
  \centering
  \includegraphics[width=0.9\linewidth]{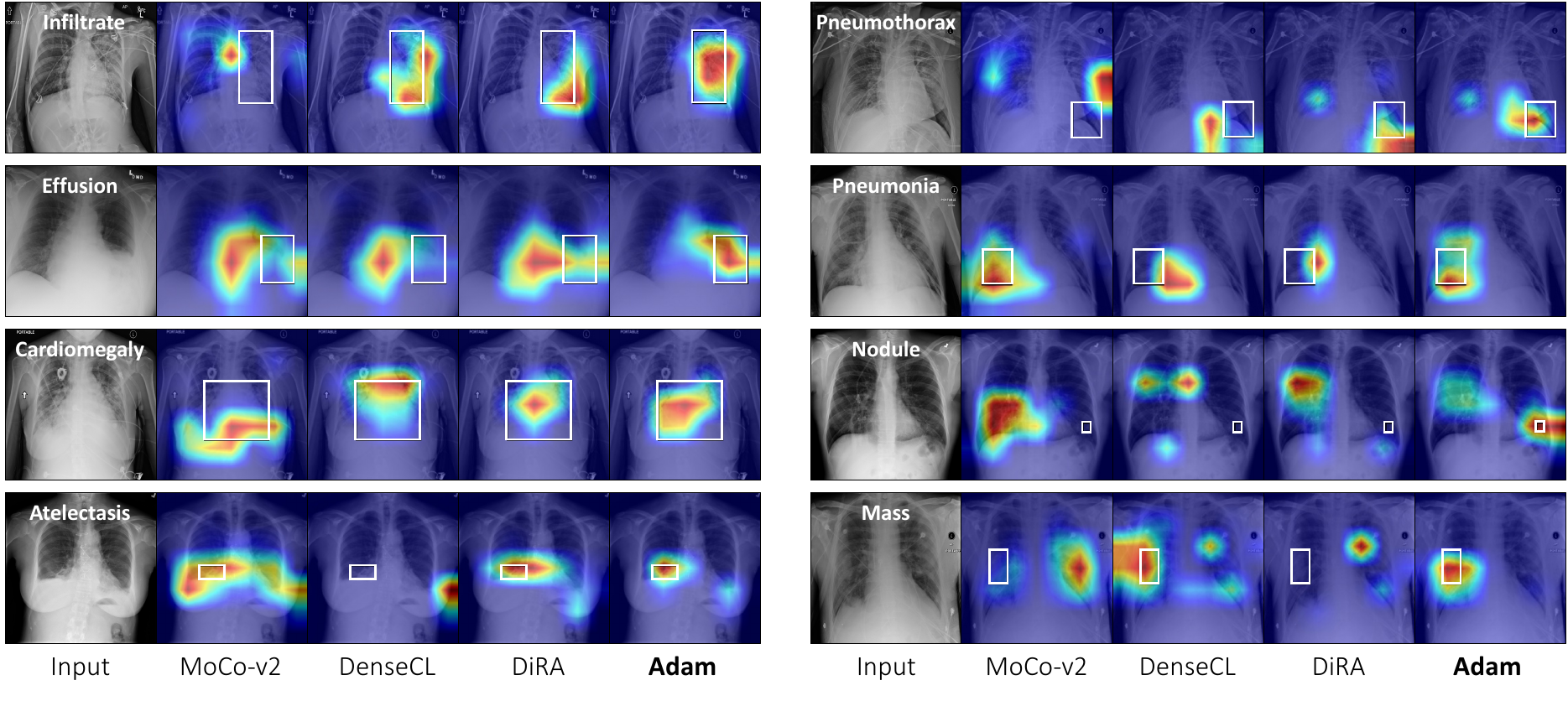}
  \caption{Visualization of Grad-CAM heatmaps generated by Adam and the bestperforming SSL methods for eight diseases in ChestX-ray14. White boxes indicate
ground truth. Adam provides more precise localization results than baselines that
focus on larger image regions or fail to overlap with the ground truth.}
 \label{fig:weakly_sup}
\end{figure}

\subsection{GradCAM visualizations for disease localization} 
We further assess the efficacy of Adam's representations for weakly-supervised disease localization. To do so, we use ChestX-ray14 dataset, which provides bounding box annotations of 8 abnormalities for around 1,000 test images. The images with bounding box annotations are only used during the testing phase to evaluate the localization accuracy. For training, we initialize the downstream model with Adam's pretrained weights and fine-tune it using only image-level disease labels. Then, following~\cite{wang2017chestx}, we calculate heatmaps using GradCAM to approximate the spatial location of a particular disease. We compare Adam with the best performing SSL methods from each baseline group (i.e. instance-level, patch-level, and
pixel-level). \cref{fig:weakly_sup} shows examples of GradCAM for Adam and other SSL baselines in eight thoracic diseases, including {\em  Atelectasis, Cardiomegaly, Effusion, Infiltrate, Mass, Nodule, Pneumonia, Pneumothorax}. As seen, Adam captures the diseased areas more precisely than the baselines. In particular, SSL baselines’ attention maps either focus on larger image regions or don't overlap with the ground truth, whereas Adam provides more robust localization results across all diseases. These findings highlight Adam's ability to learn dense representations that are more useful for disease localization.

\begin{figure}[t]
  \centering
  \includegraphics[width=0.9\linewidth]{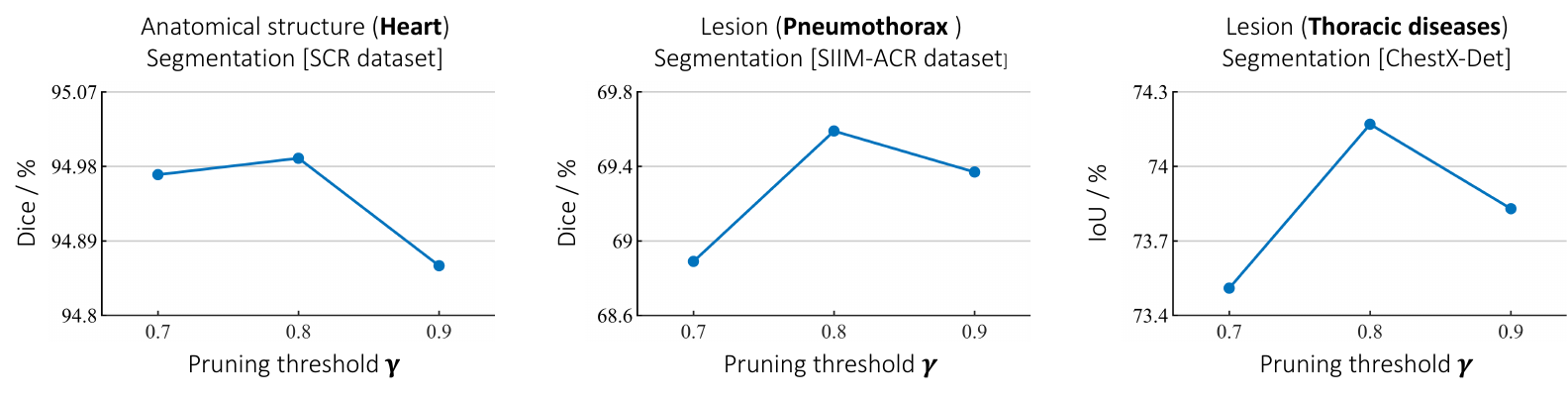}
  \caption{We conduct ablation study on the impact of pruning threshold on the downstream task performance on three downstream tasks.  The best performance achieved with $\gamma = 0.8$ in all applications.
  }
 \label{fig:gamma_ablation}
\end{figure}

\subsection{Ablation study on pruning threshold}
To explore the impact of pruning threshold ($\gamma$) of our PP module on the performance of downstream tasks, we have conducted
extensive ablation studies on different values of $\gamma$. To do so, we pretrain Adam with three pruning thresholds 0.7, 0.8, and 0.9, and transfer the pretrained model with each pruning threshold to three downstream tasks, including SCR-Heart, SIIM-ACR, and ChestX-Det. \cref{fig:gamma_ablation}  depicts the performance of Adam on three downstream tasks under different pruning thresholds. The best performance achieved at $\gamma = 0.8$ in all applications.

\begin{algorithm*}[t]
\small
   \SetAlgoLined  
  \KwIn{Anchor embeddings $q$; \newline
  Granularity level $n$; \newline
  Pruning threshold $\gamma$; \newline
  Memory bank $\mathrm{MB}$;
  }
  \KwOut{Pruned memory bank $\mathrm{MB_{pruned}}$}
\newcommand\mycommfont[1]{\footnotesize\ttfamily\textcolor{PineGreen}{#1}}
\SetCommentSty{mycommfont}

\SetKwInOut{Input}{inputs}
  \SetKwInOut{Output}{output}
    \uIf{$n=0$}{
    $\mathrm{MB_{pruned}}=\mathrm{MB}$ \;
  }
 
  \Else{
   \tcp{remove semantically similar patches to anchor from the memory bank}
   \tcp{sim(x,y) = $\frac{x}{\|x\|_2}.\frac{y}{\|y\|_2 }$ }
    \ForEach{$k_i \in \mathrm{MB}$}{
    \uIf{sim($k_i,q$) $< \gamma$}{
   $ \mathrm{MB_{pruned}} \leftarrow k_i$ \;
    }
    }
  }
   \BlankLine
 
   \caption{Purposive Pruner}
   \label{algo:purposive_pruner} 
\end{algorithm*}

\section{Purposive pruner algorithm}

Algorithm \ref{algo:purposive_pruner} presents the details of our purposive pruner (PP) component.

\section{Datasets and downstream tasks}
We pretrain Adam on two publicly available datasets, and thoroughly evaluate the transfer capability of Adam’s representations in a wide range of 9 challenging downstream tasks on 8 publicly available datasets in chest X-ray and fundus modalities. In the following, we describe the details of datasets and downstream tasks used in our study. 

\smallskip
\noindent\textbf{(1) ChestX-ray14---multi-label classification:} ChestX-ray14 dataset provides 112K chest radiographs taken from 30K unique patients, along with 14 thoracic disease labels. Each individual image may have more than one disease label. The downstream task is a multi-label classification in which the models are trained to predict 14 diseases for each image. We use the official patient-wise split released by the dataset, including 86K training images and 25K testing images. We use mean AUC over 14 diseases to evaluate the multi-label classification performance. Moreover, we use the unlabeled training data for pretraining of Adam and other self-supervised baselines.

\smallskip
\noindent\textbf{(2) NIH Shenzhen CXR---binary classification:} NIH Shenzhen CXR dataset provides 662 frontal-view chest radiographs, among which 326 images are normal and 336 images are patients with tuberculosis (TB) disease. The downstream task is a binary classification in which the models are trained to detect TB in images. We randomly divide the dataset into a training set (80\%) and a test set (20\%). We report AUC score to evaluate the classification performance.

\smallskip
\noindent\textbf{(3) VinDR-CXR---multi-label classification:} VinDR-CXR dataset provides 18,000 postero-anterior (PA) view chest radiographs that were manually annotated by a total of 17 experienced radiologists  for the classification of 5 common thoracic diseases, including pulmonary embolism, lung tumor, pneumonia, tuberculosis, and other diseases. The dataset provides an official split, including a training set of 15,000 scans and a test set of 3,000 scans. We utilize the official split, and report AUC score to evaluate the classification performance. 

\smallskip
\noindent\textbf{(4) SIIM-ACR---lesion segmentation:} SIIM-ACR  dataset provides 10K chest radiographs, including normal cases and cases with pneumothorax disease. For diseased cases, pixel-level segmentation masks are provided. The downstream task is pneumothorax segmentation. We randomly divided the dataset into training (80\%) and testing (20\%). We use mean Dice score to evaluate segmentation performance. 

\smallskip
\noindent\textbf{(5) ChestX-Det---lesion segmentation:} ChestX-Det dataset consists of 3,578 images from  ChestX-ray14 dataset. This dataset provides segmentation masks for 13 thoracic diseases, including atelectasis, calcification, cardiomegaly, consolidation, diffuse nodule, effusion, emphysema, fibrosis, fracture, mass, nodule, pleural thickening, and pneumothorax. The images are annotated by 3 board-certified radiologists. The downstream task is pixel-wise segmentation of abnormalities in images. We randomly divided the dataset into training (80\%) and testing (20\%). We use the mean IoU score to evaluate the segmentation performance. 

\smallskip
\noindent\textbf{(6) SCR-Heart\&Clavicle---organ segmentation:} SCR dataset provides 247 posterior-anterior chest radiographs from JSRT database along with segmentation masks for the heart, lungs, and clavicles. The data has been subdivided into two folds with 124 and 123 images. We follow the official split of the dataset, using fold1 for training (124 images) and fold2 for testing (123 images).  We use the mean Dice score to evaluate the heart and clavicles segmentation performances.

\smallskip
\noindent\textbf{(7) VinDR-Rib---organ segmentation:}  VinDR-Rib dataset contains 245 chest radiographs that were obtained from VinDr-CXR dataset and were manually labeled by human experts. The dataset provides segmentation annotations for 20 indivisual ribs. We use the official split released by the dataset, including a training set of 196 images and a validation set of 49 images. We use mean Dice score to evaluate segmentation performance. 

\smallskip
\noindent\textbf{(8) EyePACS---self-supervised pretraining:} EyePACS dataset consists of 88,702 colour fundus images. Expert annotations for the presence of Diabetic Retinopathy (DR) with a scale of 0–4 were provided for each image. The dataset provides an official split, including 35,126 samples for training and 53,576 samples for testing. We use unlabeled training images for self-supervised pretraining of Adam and other SSL baselines. 

\smallskip
\noindent\textbf{(9) DRIVE---organ segmentation:} The Digital Retinal Images for Vessel Extraction (DRIVE) dataset includes 40 color fundus images along with expert annotations for retinal vessel segmentation. The set of 40 images was equally divided into 20 images for the training set and 20 images for the testing set. We use the official data split and report the mean Dice score for the segmentation of blood vessels.

\section{Implementation details}
\subsection{Pretraining protocol}
In our training strategy, we use a standard ResNet-50 as the backbone in accordance with common protocol~\cite{Wang2021Dense,Kaku2021Intermediate,Haghighi2022DiRA}. Any other sophisticated backbones (i.e., variants of convolutional neural networks or vision transformers) can, however, be leveraged in our proposed training strategy. In this study, we aim to  dissect the importance of training strategy in blazing the way for learning generalizable representaitons. As such, we control other confounding factors, including the pretraining data. Consequently, Adam and all self-supervised baseline methods are pretrained on the same pretraining data from ChestX-ray14 and EyePACS datasets. We closely follow the settings of~\cite{chen2020improved} for the training parameters, including the architecture of projection heads (i.e. two-layer MLP), memory bank size (i.e. $K=65536$), contrastive temperature scaling (i.e. $\tau=0.2$), and  momentum coefficient (0.999). We use even values for $n$ and continue the training process up to $n=4$, but one can continue the training process with finer data granularity levels. It should be noted that our PP module impose negligible computational cost to the pretraining stage. We use a batch size 256 distributed across 4
Nvidia V100 GPUs with a memory of 32 GB per-card. At each training stage $n$, we train the model for 200 epochs.
 
 \subsection{Fine-tuning protocol}
 We transfer Adam's pretrained backbone (i.e., $f_\theta$) to the downstream classification tasks by appending a task-specific classification head. For the downstream segmentation tasks, we employ a U-Net network  with a ResNet-50 encoder, where the encoder is initialized with the pre-trained backbone. Following the standard protocol~\cite{Haghighi2022DiRA,Taher2021Systematic}, we evaluate the generalization of Adam’s representations by fine-tuning all the parameters of downstream models. We use input image resolution 224$\times$224 and 512$\times$512 for downstream tasks on chest X-ray and fundus images, respectively. We endeavor to optimize each downstream task with the best-performing hyperparameters as follows. For downstream classification tasks,  we use standard data augmentation techniques, including random rotation by $(-7, 7)$ degree, random crop, and random horizontal flip with probability 0.5. We follow~\cite{Xiao2023Delving} in training settings, including AdamW optimizer with weight decay 0.05, $\beta_1,\beta_2=(0.9, 0.95)$,  learning rate $2.5e-4$,  and cosine annealing learning rate decay scheduler. For downstream segmentation tasks, we use standard data augmentation techniques, including  random gamma,  elastic transformation,  random brightness contrast,  optical distortion, and  grid distortion. We use Adam optimizer with learning rate $1e-3$ for VinDR-Ribs and AdamW optimizer with a learning rate $2e-4$ for the rest of the tasks. We use cosine learning rate decay scheduler and early-stopping using  10\% of the training data as the validation set. We run each method ten times on each task and report the average, standard deviation, and statistical analysis based on an independent two-sample t-test.

 \section{Acknowledgements}
 This research has been supported in part by ASU and Mayo Clinic through a Seed Grant and an Innovation Grant, and in part by the NIH under Award Number R01HL128785. The content is solely the responsibility of the authors and does not necessarily represent the official views of the NIH. This work has utilized the GPUs provided in part by the ASU Research Computing and in part by the Bridges-2 at Pittsburgh Supercomputing Center through allocation BCS190015 and the Anvil at Purdue University through allocation MED220025 from the Advanced Cyberinfrastructure Coordination Ecosystem: Services \& Support (ACCESS) program, which is supported by National Science Foundation grants \#2138259, \#2138286, \#2138307, \#2137603, and \#2138296. The content of this paper is covered by patents pending.

\end{document}